\pdfoutput=1
\documentclass[11pt, letterpaper, onecolumn, logo, copyright]{template_from_google}

\usepackage[T1]{fontenc}
\usepackage[utf8]{inputenc}

\usepackage{amsmath}
\usepackage{amsfonts}
\usepackage{amssymb}
\usepackage{amsthm}

\usepackage{booktabs}
\usepackage{fancyhdr}
\usepackage{wrapfig}
\usepackage{authblk}
\usepackage{siunitx}
\usepackage{makecell}
\usepackage{graphicx}
\usepackage{caption}
\usepackage[dvipsnames]{xcolor}
\usepackage{framed}

\usepackage{url}

\usepackage[numbers, sort&compress, square]{natbib}

\usepackage{hyperref}
\definecolor{darkblue}{rgb}{0.0, 0.0, 0.6}
\definecolor{darkred}{rgb}{0.7, 0.0, 0.0}
\hypersetup{
  pdffitwindow=true,
  pdftitle={Conditional Residual Prediction: Improving Autoregressive Video Diffusion without a Bidirectional Teacher},
  pdfauthor={Bowen Zheng, Zhiguang Liu, Jiarong Ou, Rui Chen, Tianyang Hu},
  pdfstartview={FitH},
  pdfnewwindow=true,
  colorlinks,
  linktocpage=true,
  linkcolor=darkred,
  urlcolor=darkblue,
  citecolor=darkblue
}

\titleformat{\section}{\large\bfseries\headingfont}{\thesection.}{0.5em}{\phantomsection#1}[]
\titleformat{\subsection}{\bfseries}{\thesubsection.}{0.5em}{\phantomsection#1}[]
\titleformat{\subsubsection}{\bfseries\itshape}{\thesubsubsection.}{0.5em}{\phantomsection#1}[]

\definecolor{shadecolor}{gray}{0.96}
\theoremstyle{definition}
\newtheorem{hypothesis}{Hypothesis}
\newtheorem{definition}{Definition}
\newenvironment{hypothesisbox}[1][]
  {\begin{shaded*}\vspace*{-\topsep}\begin{hypothesis}[#1]}
  {\end{hypothesis}\vspace*{-\topsep}\end{shaded*}}

\newenvironment{definitionbox}[1][]
  {\begin{shaded*}\vspace*{-\topsep}\begin{definition}[#1]}
  {\end{definition}\vspace*{-\topsep}\end{shaded*}}

\title{Conditional Residual Prediction: Improving Autoregressive Video Diffusion without a Bidirectional Teacher}
\newcommand{\shorttitle}{Conditional Residual Prediction}
\renewcommand{\headrulewidth}{0.4pt}
\renewcommand{\footrulewidth}{0pt}
\fancypagestyleassign{firstbase}{firststyle}
\fancypagestyle{firststyle}[firstbase]{\renewcommand{\footrulewidth}{1pt}}

\reportnumber{}
\renewcommand{\today}{}

\author[1]{Bowen Zheng}
\author[2]{Zhiguang Liu}
\author[2]{Jiarong Ou}
\author[2]{Rui Chen}
\author[1]{Tianyang Hu}

\affil[1]{The Chinese University of Hong Kong, Shenzhen}
\affil[2]{Tencent Hunyuan}

\correspondingauthor{Tianyang Hu}
\renewcommand{\copyrightext}{\footerfont\itshape Emails: Bowen Zheng,
\href{mailto:bowenzheng3@link.cuhk.edu.cn}{bowenzheng3@link.cuhk.edu.cn};
Tianyang Hu, \href{mailto:hutianyang@cuhk.edu.cn}{hutianyang@cuhk.edu.cn}}

\begin{abstract}
Causal video diffusion models generate video autoregressively, which suits
streaming, interactive, and long-video generation. Under standard training,
however, they often yield lower generation quality than bidirectional models of
the same size. Many existing approaches address this gap by initializing from or
distilling a pretrained bidirectional teacher. We instead train a causal model
from an image-model initialization, with no bidirectional video model at any
stage. Because this path requires neither a large bidirectional teacher nor a
complex distillation pipeline, it is simpler and more scalable. On this path, we
find that a causal model trained on ground-truth history becomes strongly
dependent on it, so that at inference errors in its own generated history
propagate forward. We hypothesize that much of this dependence is unnecessary,
because the current input already determines much of what the history provides.
We propose Conditional Residual Prediction (CRP), a simple recipe for reducing a
model's reliance on a condition: the model first predicts the target without the
condition, and the condition may only add a residual on top of this prediction.
Applied to history, CRP makes the model predict each chunk from the present as
far as it can and use the past only for what the present cannot supply. In
controlled experiments, CRP nearly closes the 6.14-point gap to a bidirectional
model trained under the same setup. Scaling this recipe, we train Optica, a
2B-parameter causal video model that autoregressively generates 5-second 480p
videos and reaches 82.78 on VBench with only about 15M training videos.
\end{abstract}

\begin{document}

\maketitle

\section{Introduction}
Many high-quality video diffusion systems are fixed-window, non-causal
generators: they jointly denoise all latent frames in a clip, and their temporal
operators can exchange information in both directions within that window~\citep{ho2022video,blattmann2023videoldm,blattmann2023svd,cogvideox,
kong2024hunyuanvideo,wan2025wan,sanavideo}. Throughout this paper, ``bidirectional'' refers specifically to
that connectivity, not to the forward and reverse processes of diffusion.
Fixed-window generation is effective for short clips, but it must complete a
window before emitting output and does not expose the temporal KV-cache
structure used by streaming causal models~\citep{causvid,selfforcing,longlive,rollingforcing}. By contrast, the
autoregressive generation process of causal video models is naturally
compatible with long-video and streaming settings~\citep{causvid,skyreelsv2,magi1,selfforcing,longlive,rollingforcing},
with direct applications to practical scenarios such as world models and
neural renderers~\citep{worldmodels,genie,gamengen,diffusionrenderer}.

Despite these theoretical advantages, causal video models often remain
unsatisfactory in practice. More specifically, even for short durations such as
5 seconds, a causal video model produces videos with substantially lower visual
quality and semantic fidelity than a bidirectional model with a comparable
parameter count.

Recent works have recognized this phenomenon and proposed several solutions.
Early methods initialize a causal student from a pretrained bidirectional video
model and retain a bidirectional teacher for ODE initialization or score
distillation~\citep{causvid,selfforcing}. Later methods still start from a
bidirectional base but replace the ODE- or score-providing teacher with a causal
one, at initialization or throughout scoring~\citep{causalforcing,causalforcingpp,cmd}.

\begin{figure}[t]
\centering
\includegraphics[width=\textwidth]{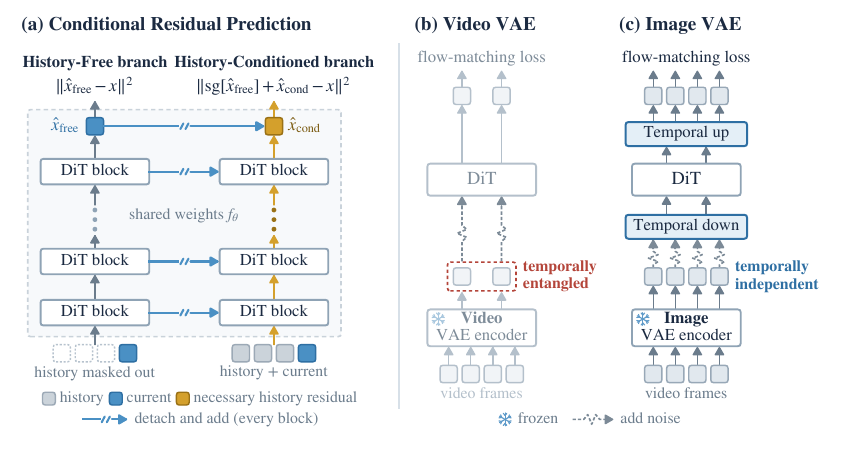}
\caption{\textbf{Overview.} \textbf{(a)} Conditional Residual Prediction (CRP)
applied to history. The same network $f_\theta$ is evaluated twice on the same
input: the History-Free branch masks out the history and predicts
$\hat{x}_{\mathrm{free}}$, and the History-Conditioned branch sees the history
and predicts only a residual $\hat{x}_{\mathrm{cond}}$ on top of it. The
History-Free hidden states are detached and added to the History-Conditioned
branch after every block, never the reverse, so the History-Free branch stays
free of history. \textbf{(b)} A video VAE encodes the first frame into its own
latent and every later four frames into one, with causal temporal convolutions
that entangle latents across time. \textbf{(c)} An image VAE encodes every frame
independently into the same kind of latent; a temporal down/up layer keeps the
cost of the DiT unchanged.}
\label{fig:crp}
\end{figure}

Compared with vanilla causal video models, these methods have achieved clear
success, approaching or even surpassing their bidirectional teacher
models. Yet their complex training and reliance on a pretrained teacher limit
scaling: a stronger causal model first needs a stronger bidirectional one. More
importantly, recent work finds
that using a causal rather than bidirectional teacher for distillation yields a
better causal student~\citep{causalforcing,cmd}: the mismatch between
bidirectional and causal models limits post-distillation performance, and a
bidirectional teacher can leak future information that the student cannot
access at generation time. This line of work further motivates our objective.
Directly training a multi-step causal video model without relying on a
bidirectional model not only provides a simple and scalable multi-step generator
in its own right, but may also enable distillation into a stronger, efficient
few-step model.

Our approach centers on one idea: reducing over-reliance on history. We find that a causal model trained on a
ground-truth prefix becomes strongly dependent on it, and we hypothesize that
much of this dependence is unnecessary: it covers information that the current
input already determines (Sec.~\ref{sec:hypothesis}). We propose Conditional Residual Prediction
(CRP; Fig.~\ref{fig:crp}(a)), which first predicts the target without a condition
and lets the condition add only a residual on top of this prediction. Applied
to history, CRP splits the forward passes over the same input into a
History-Free branch and a History-Conditioned branch and combines them to form
the final output. To prevent the History-Conditioned branch from contaminating
the History-Free branch, we detach the output of the History-Free branch with a
stop-gradient, so that the History-Conditioned loss does not backpropagate
through it; the shared parameters are updated by both losses. Under this design, the
History-Conditioned branch effectively models the residual of the History-Free
prediction. Intuitively, the History-Free branch models as much as possible
without relying on history, while the History-Conditioned branch supplies only
the signals that necessarily depend on history. In our controlled experiments,
CRP, together with the \(x_0\)-prediction it builds on, narrows the VBench
Total gap between the causal and bidirectional models from 6.14 to
0.19, with CRP itself accounting for 4.68 points.

CRP acts within the DiT, but the video VAE introduces a second source of
history dependence that no change to the DiT can remove: its causal temporal
compression couples latent frames. We therefore also use an image VAE rather
than a video VAE. On top of CRP, the image VAE further narrows the gap from
0.19 to $-0.20$. It also suits CRP: encoding every frame independently removes the
encoder-induced distinction between single-frame and four-frame latents.

Overall, this paper demonstrates that a causal video model can achieve strong
performance without relying on a bidirectional teacher: building on CRP, we
train Optica, a 2B-parameter causal video model that reaches 82.78 on VBench.
It also demonstrates the effectiveness of reducing history over-reliance and
offers a simple recipe for native, scalable training of causal video
models.

\section{Background}

Let the latent sequence of a video be \(x=(x_1,\dots,x_N)\), where the
subscript denotes a position along the temporal dimension. Let \(y\) be the text
condition and \(f_\theta\) the DiT to be trained. We use this notation
throughout the paper.

\subsection{Bidirectional Video Diffusion Model}

In image diffusion, attention is typically bidirectional~\citep{ho2020ddpm,song2021sde,rombach2022ldm,peebles2023dit}. This carried over
naturally to video diffusion, where it has been successful~\citep{ho2022video,singer2023makeavideo,ho2022imagenvideo,
blattmann2023videoldm,animatediff,blattmann2023svd} and now dominates the field~\citep{cogvideox,opensoraplan,wan2025wan,hunyuanvideo15,ltx2,seedance2,
minimaxh3}. Such methods draw one noise level \(t\) shared
across all positions, form the interpolation
\(x^{t} = (1-t)\,x + t\,\epsilon\) with \(\epsilon\sim\mathcal{N}(0,I)\), and
regress the corresponding velocity~\citep{lipman2023flow,liu2023rectifiedflow}. Attention is bidirectional along
time, so every position attends to every other within the window, and
generation integrates the ODE jointly over the whole window from \(t=1\),
producing all positions at once.

\subsection{Causal Video Diffusion Model}\label{sec:bg-direct}

Causal video diffusion models~\citep{artv,gamengen,
diffusionforcing,ruhe2024rollingdiffusion,pyramidflow,ardiffusion,skyreelsv2,
magi1,far,nova} differ from bidirectional models above all in how they treat
time, factorizing it by the chain rule:
\(p(x \mid y) = \prod_{i=1}^{N} p(x_i \mid x_{<i},\, y)\). A matching causal
attention mask in the DiT keeps information flowing in one direction only. Some
works apply causality at the chunk level~\citep{causvid,selfforcing,longlive,magi1} rather than the frame level~\citep{artv,gamengen,diffusionforcing,ardiffusion,far,nfd,scd},
which places them between the two extremes; adjusting the chunk size turns such
a model into either a bidirectional or a causal one~\citep{flexforcing}.
Throughout this paper we call a model causal when its chunk size is below two
seconds and bidirectional when it is above.

Training normally uses teacher forcing~\citep{williams1989learning}: the model
is conditioned on the ground-truth prefix and noise is added only to the current
position, so the noised query at position \(i\) attends to the clean prefix
\(x_{<i}\) and to itself:
\(\mathcal{L}_{\mathrm{causal}} = \mathbb{E}[\sum_{i=1}^{N}
\| f_\theta(x_i^{t_i},\, x_{<i},\, t_i,\, y) - (\epsilon_i - x_i) \|^2]\).
Here, each position has an independent noise level \(t_i\), while the
conditioning prefix \(x_{<i}\) comes from real data. Inference instead proceeds
autoregressively, that is, on the model's own generated frames as history: for
\(i=1,\dots,N\), the model conditions on the generated prefix \(\hat{x}_{<i}\)
and integrates the ODE from \(x_i^{1}\sim\mathcal{N}(0,I)\) to obtain
\(\hat{x}_i\). Under autoregressive inference the generated history may differ
from the real history, which introduces a training--inference gap and hence a loss
of quality. This is the central problem studied in this paper.

\subsection{Initialization and Distillation from a Bidirectional Teacher}\label{sec:bg-distill}

Many works address the quality degradation of causal video models by exploiting
a high-quality bidirectional model: some build on it by initialization or
fine-tuning, as short-chunk causal models~\mbox{\citep{ca2vdm,framepack}} or clip- and
shot-level extenders~\citep{streamingt2v,lct,svi},
while others distill from it, usually with Diff-Instruct~\citep{diffinstruct} or
DMD~\citep{dmd} objectives~\citep{causvid,selfforcing,selfforcingpp,longlive,
rollingforcing,contextforcing,causalrcm,scd}, or post-train the initialized
causal model adversarially~\citep{aapt}. This line of work has been markedly successful, with the
causal student approaching or even surpassing its bidirectional teacher.

Recent work further finds that the bidirectional teacher is itself a
bottleneck, and that making each stage causal works better. Self Forcing~\citep{selfforcing} replaces the ground-truth prefix with the model's own
rollout during training. Causal Forcing~\citep{causalforcing} traces the
problem to ODE initialization and uses a teacher-forced multi-step causal model
as the ODE teacher. Context-Matched Distillation~\citep{cmd} makes the score-providing teacher causal as well, since a
bidirectional one sees future frames the student cannot.
Appendix~\ref{app:related} expands these comparisons.

\section{Method}
\begin{figure*}[t]
  \centering
  \includegraphics[width=\textwidth]{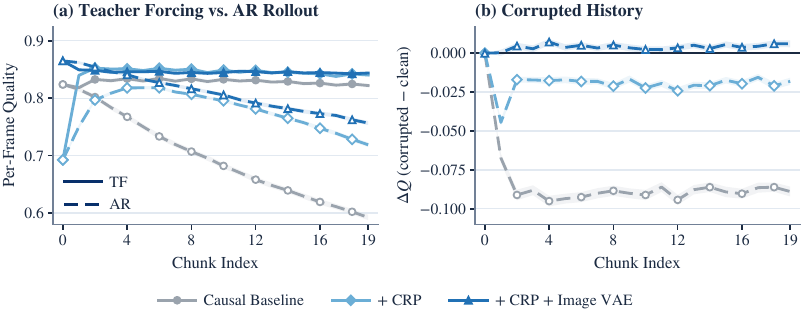}
  \caption{\textbf{Per-frame image quality degradation and comparison.}
  Per-frame no-reference quality is measured with Q-ReAlign Mini~\citep{wu2024qalign}. \textbf{(a)} Teacher-forced (TF) generation remains
  approximately stable, whereas autoregressive (AR) quality decreases with
  chunk index. \textbf{(b)} Quality change under a corrupted history,
  $\Delta Q=Q_{\mathrm{corrupted}}-Q_{\mathrm{clean}}$, with each corruption
  type triggered independently per history frame with probability $0.3$;
  negative values indicate degradation. Curves show means and shaded regions
  standard errors across clips. CRP and the image VAE progressively reduce
  the degradation. In (a), with the video VAE, $+$ CRP falls below the baseline at
  chunk 0, which is generated without history; the image VAE removes this drop
  (Sec.~\ref{sec:vae}).}
  \label{fig:history-quality}
  \end{figure*}
\subsection{The History Over-Reliance Hypothesis}\label{sec:hypothesis}
The most important difference between a causal video model and a bidirectional
model is the former's autoregressive generation process: the model first
generates the earlier portion and then uses it as the basis for generating later
positions. A common source of poor quality during autoregressive generation is
exposure bias~\citep{bengio2015scheduled,lamb2016professor,
ross2011reduction,ranzato2016sequence}. Errors in early positions are not only
carried forward as fixed history and propagated to later positions, but also
move the sequence away from the ground-truth history distribution seen during
training, thereby degrading model performance. This train--test mismatch has
also been observed directly in autoregressive video diffusion~\citep{selfforcing,resamplingforcing}. We first quantify the mismatch through per-frame
image quality, scored with Q-ReAlign Mini\footnote{\url{https://huggingface.co/q-future/Q-ReAlign-Mini-0.8B}}, a modernized version of
Q-Align~\citep{wu2024qalign}, over
$1{,}000$ real clips (Appendix~\ref{app:eval-quality}). On our causal baseline
(Sec.~\ref{sec:baseline}), quality under a ground-truth prefix is flat across
the clip, $0.824$ in the first chunk and $0.822$ in the twentieth, whereas the
same model under an autoregressive rollout falls to $0.592$, a mean gap of
$0.138\pm0.003$ (Fig.~\ref{fig:history-quality}(a)). Corrupting the
ground-truth history rather than replacing it with generated frames reproduces
the effect, costing $0.089\pm0.001$ (Fig.~\ref{fig:history-quality}(b)). We use
these two diagnostics throughout to test whether an intervention reduces
history reliance, independently of VBench.

History dependence thus poses an apparent dilemma. The model must rely on the
history to generate the current position, yet this reliance lets errors
accumulate. Simply weakening or severing its access to the history reduces
these errors but also removes useful signals, which hurts performance. We
decompose this problem from an information-theoretic perspective. Information
provided by history may be redundant with information at later positions; that
is, some information can be recovered from the current-position signal alone,
without relying on history.

More precisely, let the history be \(H=x_{<i}\), the current input be
\(c=x_i^t\), the target be \(x=x_i\), and the text input be \(y\), where \(i\)
denotes the temporal position and \(t\) the noise level in the diffusion or
flow-matching formulation. By the chain rule for mutual information~\citep{cover2006elements},
\[
I(x;c,H,y) = I(x;c,y) + I(x;H\mid c,y).
\]
We call the first term the \emph{history-free} information, which the current
input and text determine on their own, and the second the
\emph{history-exclusive} information, which only the past can supply.

\begin{hypothesisbox}[History Over-Reliance]\label{hyp:overreliance}
A causal video model trained on a ground-truth prefix relies on the history for
history-free information instead of modeling it from the current input, and
thereby acquires history dependence that the task does not require.
\end{hypothesisbox}

Why is this hypothesis plausible, and why would it cause severe exposure bias?
During training, the model receives a ground-truth prefix, so relying on the
ground-truth prefix for this history-free signal yields a lower error than
generating it independently. At inference, however, the history itself contains
errors that accumulate during generation, exacerbating exposure bias. If the model generates this component by itself,
the training loss may be higher, but at inference time the component no longer
depends on history and therefore avoids interacting with its accumulated
errors.

We therefore separate history-free and history-exclusive information
explicitly: the former is modeled from the current input, and a
History-Conditioned branch models only the remaining residual. We formalize this design as Conditional Residual
Prediction in Sec.~\ref{sec:branch}, and complement it in Sec.~\ref{sec:vae}
by removing the history dependence that the video VAE introduces.
CRP also differs from two standard techniques: training on corrupted history and
on rollout history. Diffusion Forcing~\citep{diffusionforcing}, the
best-known example of the former, injects noise into the history during
training. If inference uses a clean history, training and inference become
inconsistent; if inference keeps the noise, it discards history information
that the current position needs, by an amount set by a hand-crafted noise level
that is hard to control. Training with rollout history, or scheduled sampling~\citep{bengio2015scheduled}, changes the training objective and can lead to a
suboptimal solution~\citep{huszar2015not}. See Appendix~\ref{app:related} for
more discussion.

\subsection{Conditional Residual Prediction}\label{sec:branch}
We first focus on the DiT, where history may only add a residual on top of a
History-Free prediction.

\begin{definitionbox}[Conditional Residual Prediction]\label{def:crp}
Let a network \(f_\theta\) predict a target \(x\) from an input \(c\) and a
condition \(h\). CRP evaluates the same network twice: a condition-free branch
\(\hat{x}_{\mathrm{free}} = f_\theta(c)\) that never accesses \(h\), and a
residual branch \(\hat{x}_{\mathrm{cond}} = f_\theta(c, h)\) that may also
receive the detached features of the condition-free branch but passes nothing
back to it. With a loss \(\ell\), the two branches are trained with
\[
\mathcal{L}_{\mathrm{free}} = \ell(\hat{x}_{\mathrm{free}}, x),
\qquad
\mathcal{L}_{\mathrm{cond}} = \ell(\mathrm{sg}[\hat{x}_{\mathrm{free}}] + \hat{x}_{\mathrm{cond}}, x),
\]
where the stop-gradient \(\mathrm{sg}[\cdot]\), which we use throughout, keeps
\(\mathcal{L}_{\mathrm{cond}}\) from backpropagating through the condition-free
branch, and the
prediction is \(\hat{x}_{\mathrm{free}} + \hat{x}_{\mathrm{cond}}\), or
\(\hat{x}_{\mathrm{free}}\) alone when \(h\) is unavailable. The condition-free
branch thus carries no information about \(h\) in its features or output;
everything that depends on \(h\) is carried by the residual branch, without
additional parameters.
\end{definitionbox}

\begin{wraptable}{r}{0.48\textwidth}
  \vspace{-14pt}
  \setlength{\abovecaptionskip}{0pt}
  \caption{Roadmap of the controlled ablation, trained on 4M videos at 256p and
  5s. Each row adds one configuration change on top of the row above, and
  $\Delta$ is the resulting change in Total. The causal baseline is substantially
  worse than the bidirectional baseline, CRP nearly closes this gap, and the
  image VAE further brings it on par with the bidirectional model. All
  scores use the standard VBench prompt set and its fixed seeds;
  Appendix~\ref{app:vae-recon} compares reconstruction in image and video modes.}
  \label{tab:ladder}
  \vspace{5pt}
  \centering
  \scriptsize
  \setlength{\tabcolsep}{4pt}
  \begin{tabular}{@{}l S[table-format=2.2] S[table-format=2.2] S[table-format=2.2]
    S[table-format=+1.2,print-implicit-plus]@{}}
  \toprule
  {Configuration} & {Total} & {Quality} & {Semantic} & {$\Delta$} \\
  \midrule
  Bidirectional reference       & 77.28 & 79.64 & 67.82 & {--} \\
  \midrule
  Causal baseline               & 71.14 & 75.13 & 55.16 & {--} \\
  \quad $+$ \(x_0\)-prediction  & 72.41 & 74.96 & 62.22 & 1.27 \\
  \quad $+$ CRP                 & 77.09 & 78.09 & 73.10 & 4.68 \\
  \quad $+$ Image VAE           & 77.48 & 78.43 & 73.72 & 0.39 \\
  \bottomrule
  \end{tabular}
    \vspace{-14.5pt}
\end{wraptable}
In this paper, \(\ell\) is the squared error, and we instantiate
Definition~\ref{def:crp} with the noisy current chunk and the text as the input,
\(c=(x_i^t, y)\), the history as the condition, \(h=H=x_{<i}\), and the clean
chunk as the target, \(x=x_i\).

Conditional residual branches have conceptual precedents in ControlNet and
temporal-residual video diffusion~\citep{zhang2023controlnet,zhang2024trip}.
ControlNet, however, injects an additional, separately parameterized condition
branch into a frozen base, so that the features and output of the base carry
the condition. CRP reverses this direction within a single network and keeps the
condition-free branch clean. SCD separates temporal causal reasoning from
frame-wise denoising~\citep{scd}. Appendix~\ref{app:related} further discusses
History Guidance and remedies for over-reliance on history in imitation
learning.

Specifically, for the same network and the same set of inputs, we compute two
forward-pass results simultaneously (Fig.~\ref{fig:crp}(a)): a History-Free branch
with past frames masked out and a History-Conditioned residual branch with
access to past frames. During sampling, the first chunk, which has no history,
uses only \(\hat{x}_{\mathrm{free}}\). We pass the detached hidden state of the
History-Free branch to the History-Conditioned branch after every block,
allowing the latter to process History-Free signals in latent space rather than
only in output space; we refer to this as latent fusion. Since the residual is most naturally defined on the clean
target, CRP uses \(x_0\)-prediction.

\subsection{From Video VAE to Image VAE}\label{sec:vae}

CRP acts within the DiT, but the latents it models come from the VAE. Like many
common video VAEs, the Wan2.1 VAE we adopt~\citep{wan2025wan} is built from
causal temporal convolutions: it encodes the first frame into its own latent and
every later four frames into one, and each latent depends on the frames before
it (Fig.~\ref{fig:crp}(b)). This causes two problems.

First, the VAE introduces history dependence of its own, which no change to the
DiT can remove: it learns to reconstruct later positions partly from preceding
signals. We observe that, under the common setting of 5 seconds at 16 fps, an error in an
early latent propagates all the way to the last frame
(Appendix~\ref{app:video-vae}). In autoregressive generation, an error in an
earlier chunk is thus carried into every later one.

Second, the video VAE produces two types of latent, a single-frame latent for
the first frame and a four-frame latent for each later group of four frames, and
one model must generate both. The two are also inherently imbalanced: the
single-frame latent appears only at the first position and receives a small
share of training, which an extra image task can alleviate. In CRP, the
imbalance becomes more pronounced: the History-Free branch predicts without
history at every position, almost always on four-frame latents. Since the
first chunk is generated by this branch alone, it falls below the baseline
(Fig.~\ref{fig:history-quality}(a), chunk 0). Rather than adding an
image task, we unify the latent representation.

Motivated by these two issues, we replace the video VAE with an image VAE, which
encodes every frame
independently into the same kind of latent (Fig.~\ref{fig:crp}(c)). To avoid
introducing other changes, we keep the same Wan2.1 VAE but use its image mode
instead of its video mode, which loses no fidelity ($1.11$ dB higher paired
PSNR; Appendix~\ref{app:vae-recon}). This removes both problems: a
perturbation of one latent no longer spreads to later frames, and the chunk-0
drop disappears.

To keep the DiT's cost unchanged with four times as many latents, we add a
lightweight temporal patch layer that merges every four adjacent latents at the
input and splits them at the output (Fig.~\ref{fig:crp}(c);
Appendix~\ref{app:patch}). Consistent with JiT~\citep{li2026jit}, we observe that
\(x_0\)-prediction performs better than velocity prediction with these
higher-dimensional tokens. We refer to this
complete configuration as ``$+$ Image VAE.''

\section{Controlled Ablation Experiments}

\subsection{Setup}\label{sec:setup}
\paragraph{Data} For the controlled ablation experiments, we use approximately
4M internal video clips at 256p resolution and a duration of 5 seconds. Further
details on the data are provided in Appendix~\ref{app:data}.

\paragraph{Model} Our model is initialized from the 1.6B-parameter SANA-1.5
T2I checkpoint~\citep{sana,sana15}. Following SANA-Video~\citep{sanavideo}, we add
a temporal convolutional layer to the MLP and adapt kernelized linear attention~\citep{katharopoulos2020linear} for RoPE~\citep{su2021roformer}. We replace 25\% of the linear-attention
layers with softmax attention. For the causal
model, all components involving the temporal dimension, including temporal
convolution, attention, and RoPE, are made strictly causal. The resulting model
contains 2B parameters. We use the Wan2.1 VAE. Since SANA-1.5 is
pretrained in the DC-AE latent space~\citep{dcae}, we follow DC-VideoGen~\citep{dcvideogen} and adapt its input and output layers to the Wan2.1 latent
space before the main training (Appendix~\ref{app:vae-adapt}).

\subsection{Bidirectional and Causal Baselines}\label{sec:baseline}
\begin{figure*}[t]
\centering
\newcommand{\qualw}{0.164\textwidth}
\newcommand{\qualtime}[1]{\footnotesize #1}
\setlength{\tabcolsep}{0.6pt}
\begin{tabular}{@{}cccccc@{}}
  \includegraphics[width=\qualw]{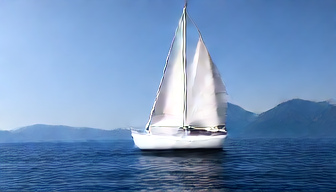}
  & \includegraphics[width=\qualw]{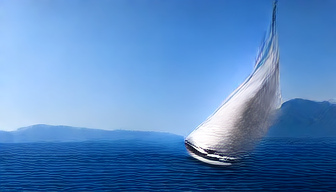}
  & \includegraphics[width=\qualw]{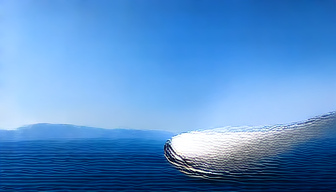}
  & \includegraphics[width=\qualw]{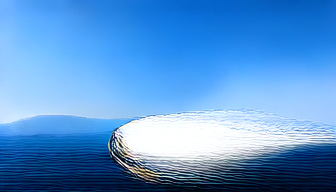}
  & \includegraphics[width=\qualw]{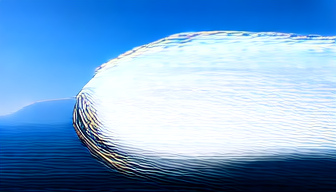}
  & \includegraphics[width=\qualw]{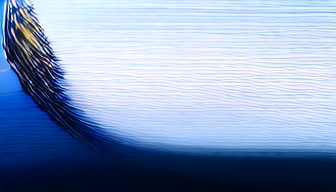} \\[0.7pt]
  \includegraphics[width=\qualw]{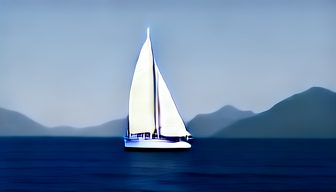}
  & \includegraphics[width=\qualw]{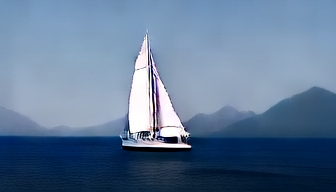}
  & \includegraphics[width=\qualw]{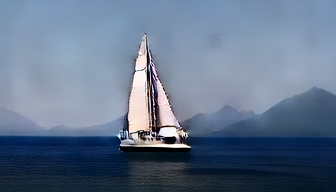}
  & \includegraphics[width=\qualw]{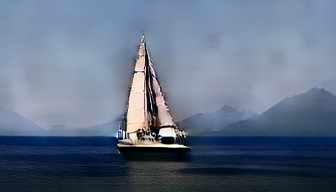}
  & \includegraphics[width=\qualw]{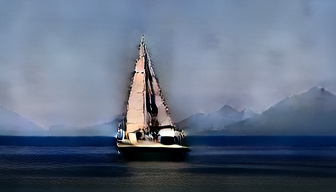}
  & \includegraphics[width=\qualw]{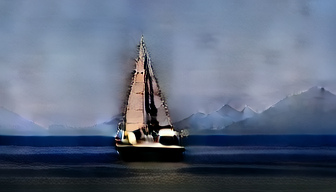} \\[0.7pt]
  \includegraphics[width=\qualw]{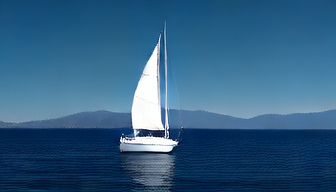}
  & \includegraphics[width=\qualw]{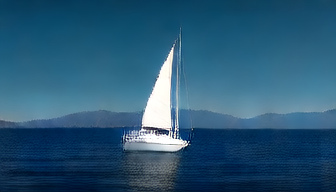}
  & \includegraphics[width=\qualw]{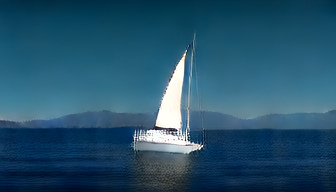}
  & \includegraphics[width=\qualw]{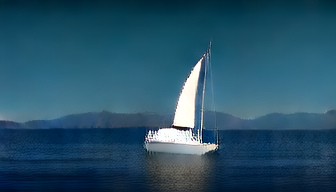}
  & \includegraphics[width=\qualw]{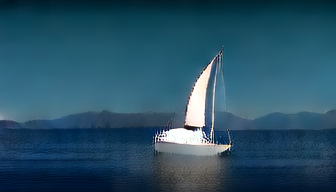}
  & \includegraphics[width=\qualw]{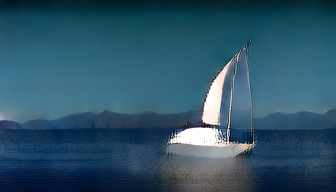} \\[1pt]
  \qualtime{$t{=}0$} & \qualtime{$t{=}16$} & \qualtime{$t{=}32$}
  & \qualtime{$t{=}48$} & \qualtime{$t{=}60$} & \qualtime{$t{=}72$} \\
\end{tabular}
\caption{\textbf{Qualitative effect of reducing history over-reliance over an
autoregressive rollout.} Rows, top to bottom: causal baseline; $+$ CRP;
$+$ CRP $+$ Image VAE. All three are the 4M-video controlled-ablation models
under the same sampling parameters, so the differences between them isolate the effect of each added component. The
causal baseline deteriorates rapidly as the temporal index increases. With $+$ CRP, the
scene keeps its structure throughout, but later frames darken and develop
speckle artifacts; adding the image VAE removes them.}
\label{fig:qualitative-ablation}
\end{figure*}

We first train bidirectional and causal baselines under this setup. For the
baseline models, we use the standard velocity loss. For the causal baseline,
we use teacher forcing rather than diffusion forcing, following prior work~\citep{causalforcing}. The causal baseline scores 71.14 VBench Total, well
below the bidirectional baseline's 77.28. This is consistent with prior
autoregressive video diffusion results~\citep{causvid,selfforcing}.
Qualitatively, we observe clear exposure bias: as
generation proceeds, the causal baseline rapidly departs from the real-image
distribution, producing distorted textures, incorrect structures, and
artifacts (Fig.~\ref{fig:qualitative-ablation}, first row).

\subsection{Causal Model with CRP}\label{sec:hf-branch}
Starting from the causal baseline, we first add CRP to the DiT, keeping the
video VAE. Because CRP predicts a residual on top of an \(x_0\) prediction, we
first switch the causal baseline from velocity prediction to
\(x_0\)-prediction, which on its own raises VBench Total from 71.14 to 72.41
(Table~\ref{tab:ladder}), and then add CRP (Sec.~\ref{sec:branch}).

CRP raises VBench Total from 72.41 to 77.09, close to the
bidirectional baseline of 77.28. CRP alone therefore accounts for 4.68
of the 6.14-point gap between the causal and bidirectional baselines.
On the two diagnostics in Fig.~\ref{fig:history-quality}, CRP reduces the
quality gap between autoregressive generation and teacher forcing from
\(0.138\pm0.003\) to \(0.066\pm0.003\), and the quality loss caused by corrupted
history from \(0.089\pm0.001\) to \(0.020\pm0.001\). In the rollout of
Fig.~\ref{fig:qualitative-ablation}, the scene keeps its structure throughout,
although later frames darken and develop speckle artifacts.

\subsection{Causal Model with Image VAE}\label{sec:image-vae}
We then replace the video VAE with the image VAE of Sec.~\ref{sec:vae},
together with the temporal patch layer, which adds about 40M parameters, or 2\%
of the total (Fig.~\ref{fig:crp}(c)). 

The image VAE further raises VBench Total from 77.09 to 77.48,
matching the bidirectional baseline. Adding the image VAE also reduces the two diagnostics to
\(0.039\pm0.002\) and \(-0.004\pm0.001\), respectively; at this corruption
strength, there is no longer any measurable degradation. It also removes the
drop of CRP at chunk 0 (Fig.~\ref{fig:history-quality}(a); Sec.~\ref{sec:vae})
and the darkening and speckles in later frames
(Fig.~\ref{fig:qualitative-ablation}).

\subsection{Ablation of CRP Components}\label{sec:crp-ablation}

\begin{table}[t]
\setlength{\abovecaptionskip}{8pt}
\caption{Ablation of the CRP components in the controlled setting. Full CRP is
the $+$ CRP row of Table~\ref{tab:ladder},
and each row below it removes the components marked $\times$. Extra History-Free
training applies \(\mathcal{L}_{\mathrm{free}}\) to every chunk, whereas the
causal baseline predicts without history only for the first chunk. Latent fusion
passes the detached hidden states of the History-Free branch to the
History-Conditioned branch after every block. CRP composes the final prediction
as \(\mathrm{sg}[\hat{x}_{\mathrm{free}}] + \hat{x}_{\mathrm{cond}}\); without
it, the History-Conditioned branch predicts \(x_0\) directly. The stop-gradient
acts on whichever of these two paths is present and does not apply ({--}) when
neither is.}
\label{tab:crp-ablation}
\centering
\small
\setlength{\tabcolsep}{7pt}
\begin{tabular}{@{}l cccc S[table-format=2.2] S[table-format=2.2] S[table-format=2.2]@{}}
\toprule
 & {\makecell{Extra History-Free\\Training}} & {\makecell{Latent\\Fusion}} & {CRP}
 & {\makecell{Stop\\Gradient}} & {Total} & {Quality} & {Semantic} \\
\midrule
Causal baseline (\(x_0\))      & $\times$   & $\times$   & $\times$   & {--}       & 72.41 & 74.96 & 62.22 \\
\midrule
Full CRP                       & \checkmark & \checkmark & \checkmark & \checkmark & 77.09 & 78.09 & 73.10 \\
\quad $-$ Latent fusion        & \checkmark & $\times$   & \checkmark & \checkmark & 76.31 & 77.06 & 73.32 \\
\quad $-$ CRP                  & \checkmark & \checkmark & $\times$   & \checkmark & 72.63 & 75.11 & 62.74 \\
\quad $-$ Both                 & \checkmark & $\times$   & $\times$   & {--}       & 71.14 & 74.08 & 59.37 \\
\midrule
\quad $-$ Stop gradient        & \checkmark & \checkmark & \checkmark & $\times$   & 75.87 & 77.39 & 69.79 \\
\bottomrule
\end{tabular}
\end{table}

\begin{figure}[t]
\centering
\includegraphics[width=\textwidth]{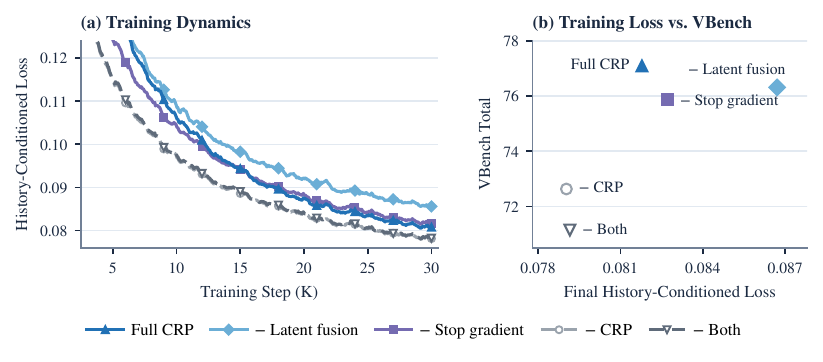}
\caption{\textbf{CRP reduces history over-reliance and improves generation.} The History-Conditioned loss is the loss of the final prediction
on chunks that have history, for the variants of Table~\ref{tab:crp-ablation}.
\textbf{(a)} Loss during training. \textbf{(b)} Final loss against VBench Total.
The variants without CRP (dashed lines, open markers) reach the lowest loss but
the lowest VBench Total.}
\label{fig:crp-ablation-loss}
\end{figure}

To isolate the contribution of each component, we remove them from the $+$ CRP
model of Sec.~\ref{sec:hf-branch} (Table~\ref{tab:crp-ablation}). Removing latent fusion lowers VBench Total from
77.09 to 76.31, and removing the stop-gradient lowers it to 75.87. Replacing the
composed prediction \(\mathrm{sg}[\hat{x}_{\mathrm{free}}] +
\hat{x}_{\mathrm{cond}}\) with a direct prediction of the History-Conditioned
branch ($-$ CRP) lowers it much further, to 72.63, close to the \(x_0\) causal baseline
of 72.41. The model therefore tends to over-rely on history: latent fusion
alone still lets it bypass the History-Free signals, and only composing the two
branches at the output forces it to model them. CRP trains \(\mathcal{L}_{\mathrm{free}}\) and \(\mathcal{L}_{\mathrm{cond}}\)
jointly, but the extra History-Free training alone does not account for its
gain. Without latent fusion and CRP ($-$ Both), VBench Total falls to 71.14,
below the \(x_0\) causal baseline of 72.41. Without CRP, the History-Free prediction is used only
for the first chunk, which has no history; we attribute the drop to the extra
History-Free training diluting the share of History-Conditioned training for
every later chunk.

Figure~\ref{fig:crp-ablation-loss} shows the History-Conditioned training loss of
these variants. Without latent fusion, the History-Conditioned branch cannot
obtain History-Free signals in latent space, and both the loss and VBench Total
are worse: the loss stays above that of full CRP throughout training and ends at
0.0867, versus 0.0818. Without the stop-gradient, the loss decreases faster at first and stays
below that of full CRP until about 16K steps, but \(\mathcal{L}_{\mathrm{cond}}\)
also updates the History-Free branch, whose loss ends 0.0012 higher; the
final loss of 0.0827 and VBench Total are both worse. Without CRP, the model
over-relies on history: its loss is lower throughout and ends at 0.0790 (0.0792
for $-$ Both), yet these two variants have the lowest VBench Total. This is the behavior that
Hypothesis~\ref{hyp:overreliance} predicts: under a ground-truth prefix,
relying on the history yields a lower training error, whereas at inference the
history contains the model's own errors (Sec.~\ref{sec:hypothesis}). Together, these results isolate the contribution of each component of CRP and
support the history over-reliance hypothesis.

\section{Scaling Experiments}\label{sec:scaling}

\begin{table}[t]
\setlength{\abovecaptionskip}{8pt}
\caption{Comparison with video generators on VBench. Total, Quality, and Semantic are short-video
VBench-T2V scores~\citep{huang2024vbench}. For our model, we report scores
on 480p$+$5s videos generated autoregressively in 5 chunks.
Appendix~\ref{app:eval-vbench} lists the sources of the quoted scores.}
\label{tab:scaling}
\centering
\footnotesize
\setlength{\tabcolsep}{3pt}
\renewcommand{\arraystretch}{0.95}
\begin{tabular}{@{}l c c l l
  S[table-format=2.2] S[table-format=2.2] S[table-format=2.2]@{}}
\toprule
Method & Steps & \#Params & Bi.\ source & Use & {Total} & {Quality} & {Semantic} \\
\midrule
\multicolumn{8}{l}{\emph{Fixed-window, non-causal}} \\
Step-Video~\citep{stepvideo}              & 50 & 30B   & --         & --             & 81.83 & 84.46 & 71.28 \\
Open-Sora-2.0~\citep{opensora2}           & 50 & 11B   & --         & --             & 84.34 & 85.40 & 80.12 \\
HunyuanVideo~\citep{kong2024hunyuanvideo} & 50 & 13B   & --         & --             & 83.24 & 85.09 & 75.82 \\
Wan2.1~\citep{wan2025wan}                 & 50 & 14B   & --         & --             & 83.69 & 85.59 & 76.11 \\
CogVideoX1.5~\citep{cogvideox}            & 50 & 5B    & --         & --             & 82.17 & 82.78 & 79.76 \\
SANA-Video~\citep{sanavideo}              & 50 & 2B    & --         & --             & 83.71 & 84.35 & 81.35 \\
LTX-Video~\citep{ltxvideo}                & 20 & 1.9B  & --         & --             & 80.00 & 82.30 & 70.79 \\
Wan2.1~\citep{wan2025wan}                 & 50 & 1.3B  & --         & --             & 84.26 & 85.30 & 80.09 \\
\midrule
\multicolumn{8}{l}{\emph{Causal, distilled from a bidirectional teacher}} \\
CausVid~\citep{causvid}                   & 4  & 1.3B  & Wan-1.3/14B & init.+dist. & 81.20 & 84.05 & 69.80 \\
Self Forcing~\citep{selfforcing}          & 4  & 1.3B  & Wan-1.3/14B & init.+dist. & 84.31 & 85.07 & 81.28 \\
LongLive~\citep{longlive}                 & 4  & 1.3B  & Wan-1.3/14B & init.+dist. & 84.87 & 86.97 & 76.47 \\
Rolling Forcing~\citep{rollingforcing}    & 5  & 1.3B  & Wan-1.3/14B & init.+dist. & 81.22 & 84.08 & 69.78 \\
Context Forcing~\citep{contextforcing}    & 4  & 1.3B  & Wan-1.3B   & init.+dist. & 83.44 & 84.98 & 77.29 \\
Causal Forcing~\citep{causalforcing}      & 4  & 1.3B  & Wan-1.3/14B & init.+dist. & 84.04 & 84.59 & 81.84 \\
Causal-rCM~\citep{causalrcm}              & 2  & 1.3B  & Wan-1.3/14B & init.+dist. & 84.63 & 85.46 & 81.31 \\
SCD~\citep{scd}                           & 4  & 1.6B  & Wan-1.3/14B & init.+dist. & 84.03 & 85.14 & 79.60 \\
\midrule
\multicolumn{8}{l}{\emph{Causal, bidirectional initialization only}} \\
SkyReels-V2~\citep{skyreelsv2}            & 30 & 1.3B  & own bi.\ model & init.       & 82.67 & 84.70 & 74.53 \\
\midrule
\multicolumn{8}{l}{\emph{Causal, no bidirectional video model}} \\
NOVA~\citep{nova}                         & 25 & 0.6B  & --         & --             & 80.12 & 80.39 & 79.05 \\
Pyramid Flow~\citep{pyramidflow}          & 20 & 2B    & --         & --             & 81.72 & 84.74 & 69.62 \\
MAGI-1~\citep{magi1}                      & 64 & 4.5B  & --         & --             & 79.18 & 82.04 & 67.74 \\
Optica (ours)                             & 50 & 2B    & --         & --             & 82.78 & 83.02 & 81.83 \\
\bottomrule
\end{tabular}
\end{table}

\paragraph{Data} Building on the controlled experiments, we continue training
at 256p and 5 seconds up to 7M videos, and then switch to 480p and train up to 14.5M videos.
Finally, we perform supervised fine-tuning (SFT) on 0.5M internal
high-quality videos (Appendix~\ref{app:data}), for 15M videos in total.

\paragraph{Model} The scaled-up model continues training from the
controlled-experiment checkpoint. All controlled
experiments use a chunk size of one latent position; for the scaled-up model,
we increase the chunk size to four latent positions and fine-tune directly.

\paragraph{Results} Optica achieves VBench
Total/Quality/Semantic scores of 82.78/83.02/81.83 (Table~\ref{tab:scaling}). It is trained on only approximately 15M internal real video clips, contains
2B parameters, and uses no initialization, ODE sample, or score from a
bidirectional teacher.

Among methods that do not rely on any bidirectional video model, Optica
achieves the highest VBench Total score: 82.78, exceeding Pyramid Flow, NOVA,
and MAGI-1 by 1.06, 2.66, and 3.60 points, respectively. This advantage comes
primarily from Semantic: Optica scores 81.83, compared with 69.62, 79.05, and
67.74. Our Quality score remains below that of Pyramid Flow (83.02 vs.\ 84.74).

We observe a similar pattern when comparing against causal models distilled
from bidirectional teachers. Optica's Semantic score is already comparable to
those of the strongest models in this group. For example, Causal Forcing,
Causal-rCM, and Self Forcing achieve Semantic scores of 81.84, 81.31, and
81.28, respectively, compared with 81.83 for Optica. On Quality, however,
Optica remains 1.0--4.0 points lower. The same holds relative to the
bidirectional SANA-Video 5-second checkpoint, the closest published model
architecturally: their Semantic scores are nearly identical (81.83 vs.\ 81.35),
and the primary gap lies in Quality (83.02 vs.\ 84.35).

The remaining gap should be read together with the training budget. Although many distilled causal models contain only 1.3B
parameters, they are initialized from or distilled with pretrained Wan video
models, commonly at 1.3B and in most cases with 14B teacher supervision~\citep{causvid,selfforcing,longlive,causalrcm,scd}. The cost of this upstream
training is not reflected in their parameter counts. By contrast, Optica uses
only about 15M video clips and starts from an image model. More importantly, in our controlled
experiments with identical data, model sizes, and evaluation protocols, our
causal model already matches the bidirectional reference
(Table~\ref{tab:ladder}). This points to training budget, rather than the
causal architecture, as the main source of the remaining gaps.

\section{Conclusion}
\label{sec:conclusion}

This paper studies history over-reliance in causal video models: teacher
forcing supplies a consistently correct prefix, so the model comes to rely on
history for signals it could model from the current position, and this reliance
amplifies history errors during autoregressive inference. We address this with
Conditional Residual Prediction, which splits the DiT into a History-Free
branch and a History-Conditioned branch that predicts only a residual on top of
it. We complement it with an image VAE that removes history dependence between
latents. In controlled experiments, CRP narrows the 6.14-point VBench
Total gap to the bidirectional reference to 0.19, and the image VAE
narrows it further to $-0.20$, while substantially reducing quality
degradation under generated and corrupted histories. The scaled
2B-parameter Optica uses no initialization, sample, or score from a
bidirectional teacher and achieves 82.78 on VBench.

History remains necessary for modeling motion, identity, and temporal
consistency; what CRP removes is the reliance on history for what the current
position can model on its own. This also makes Optica
complementary to distillation rather than an alternative to it. Recent work
finds that a few-step causal model benefits from distilling a multi-step causal
teacher rather than a bidirectional one~\citep{causalforcing}, and Optica offers a way to
train such a teacher natively, keeping the causal factorization consistent
between the pretrained model and the downstream student.

The same decomposition also suggests a target for compressing long context or
memory in causal video models: keep the history-exclusive information, which the
current position cannot infer on its own, and discard the rest. This gives an
information-theoretic rather than heuristic criterion for which parts of the
history are useful, and can guide the design of concrete compression algorithms.

Our experiments have several limitations. All controlled and scaling runs
evaluate only 5-second videos, so we have not verified that CRP
keeps suppressing error accumulation over longer autoregressive rollouts. We
also instantiate CRP only with the history as the condition; other conditions
that generative models are known to over-rely on, such as the conditioning
image in image-to-video generation~\citep{zhao2024leakage}, are left for future
work.
This version of Optica also does not yet support image-to-video generation. The
image VAE gives up temporal compression and the History-Free branch adds a
forward pass, so the current 50-step model does not demonstrate real-time
efficiency, and we have not yet distilled Optica into a few-step model.

\subsection*{AI Use Statement}
LLMs were used for language polishing, coding assistance, reviewing mathematical
formulas, and searching for related literature. The core method and research
were developed without AI assistance. The authors take full responsibility for
the final content of this paper.

{\small
\bibliography{references}
}

\appendix

\section{Extended Related Work}
\label{app:related}

\paragraph{Masked-Conditional Video Diffusion} MCVD~\citep{mcvd}, FDM~\citep{fdm}, and RaMViD~\citep{ramvid} can also be rolled out autoregressively
by conditioning on previously generated frames, but they are trained to
condition on arbitrary frames, including future ones, and are therefore not
causal in the sense of Sec.~\ref{sec:bg-direct}.

\paragraph{Causal Distillation} These works extend the pipeline of
Sec.~\ref{sec:bg-distill}. Self-Forcing++~\citep{selfforcingpp} re-noises long
rollouts to push self-rollout training past the teacher's window. LongLive~\citep{longlive} adds KV recaching at prompt switches and a frame-level
attention sink for real-time interactive streaming. Rolling Forcing~\citep{rollingforcing} jointly denoises a rolling window with bidirectional
intra-window attention while emitting frames sequentially. Context Forcing~\citep{contextforcing} distills from a long-context teacher and restructures
the KV cache. Causal-rCM~\citep{causalrcm} combines teacher-forcing
consistency distillation with self-forcing DMD under a forward/reverse
divergence view. SCD~\citep{scd} decouples once-per-frame causal reasoning
from frame-wise denoising. Causal Forcing++~\citep{causalforcingpp} inherits
Causal Forcing's teacher-forcing and DMD stages, replaces its causal ODE stage
with causal consistency distillation, and extends to frame-wise one- and
two-step generation.

\paragraph{Training with Perturbed History} Diffusion Forcing~\citep{diffusionforcing} samples each noise level independently, so that 16
frames with 1000 timesteps each already admit $O(10^{48})$ noise-level
combinations~\citep{ardiffusion}. Rolling Diffusion~\citep{ruhe2024rollingdiffusion}, AR-Diffusion~\citep{ardiffusion},
SkyReels-V2~\citep{skyreelsv2} and MAGI-1~\citep{magi1} restrict the schedule
to be non-decreasing toward the future. The clean prefix can also be corrupted
on top of a teacher-forced schedule. MAGI-1 injects up to $5\%$ noise into its
clean chunks~\citep{magi1}. SkyReels-V2 adds noise to the clean condition at
inference to stabilize long rollouts~\citep{skyreelsv2}. Pyramid Flow noises
the history during training while conditioning on clean frames at inference~\citep{pyramidflow}. GameNGen finds corrupting context frames critical for
long-horizon visual stability~\citep{gamengen}. In-Context Forcing~\citep{incontextforcing} argues that clean context frames leak local detail
and invite shortcuts; it adds more noise to adjacent frames and less to
distant ones while leaving the prediction target fixed, partly in service of
cross-frame parallel denoising, and studies few-step models downstream of a
distillation pipeline. Such corruption is not selective: it weakens the
history-exclusive information that the current position needs along with the
redundant history-free information. It also leaves a dilemma between training
and inference. With noise augmentation as in Diffusion Forcing, noise applied
only in training leaves inference with a clean history that training rarely
provides, whereas noise applied at inference as well discards information by an
amount set by the noise level rather than by what the current position needs.
Frame-, block-, and patch-level corruption or dropout face the same trade-off.
Our runs condition on a clean prefix throughout,
following \citet{causalforcing}, who report that plain teacher forcing usually
outperforms diffusion forcing, a finding our internal experiments reproduce.

\paragraph{Training with Rollout History} Self Forcing~\citep{selfforcing}
unrolls the student with KV caching inside the training loop and supervises
the rollout with a distribution-matching objective; the extensions above~\citep{selfforcingpp,longlive,rollingforcing,causalrcm} keep this structure. A
rollout carries no ground-truth target, so all of them draw one from the
distillation pipeline, and hence from a teacher. Resampling Forcing~\citep{resamplingforcing} does without one and is the closest teacher-free
work to ours: it shares our target of exposure bias under teacher forcing, our
stop-gradient, and our avoidance of any teacher or discriminator. It
autoregressively resamples the history with the online weights, detaches that
rollout, and asks the model to predict a clean target from the degraded
prefix, so it learns to \emph{rectify} history errors. Supervising a
self-generated history with the ground-truth continuation, as in scheduled
sampling~\citep{bengio2015scheduled}, is an improper objective in general~\citep{huszar2015not}: once the rollout drifts or collapses, the same generated
history is consistent with many real continuations, so the regression target
either averages over them or asks the model to recover information that the
history no longer contains. Resampling Forcing initializes from
pretrained Wan2.1-1.3B weights rather than from an image model, and its
history routing targets long-horizon attention cost, which we do not address.
Both families change the conditioning signal; CRP changes the
prediction target instead, at the price of one extra forward pass.

\paragraph{Controlling the History Dependency}
History Guidance~\citep{song2025historyguided} uses a CFG-style mechanism to
inject the history signal at inference time, allowing the strength of the
history signal to be controlled. It evaluates a single network with and without
history, but trains it to predict the full target in every case and combines
the predictions only at sampling time. MAGI-1~\citep{magi1} sets a CFG guidance
coefficient for the history separately from the text one, and schedules it
over the denoising steps. SCD~\citep{scd} separates the history-information
extractor from the frame-wise denoiser, making the denoiser depend on a
compressed history and thereby providing a bottleneck on the history signal.
In imitation learning, policies conditioned on observation histories are known
to over-rely on the history. This is addressed
with targeted interventions~\citep{dehaan2019causal}, with an adversarial
objective that removes information about the previous action~\citep{wen2020copycat}, or by training a separate history module to predict the
change of action and feeding its detached features to a policy that
predicts the full action~\citep{chuang2022copycat}. By contrast, the residual in
CRP is added to a detached History-Free prediction of the same network.

\paragraph{Image-Initialized Causal Video Models} Pyramid Flow~\citep{pyramidflow} initializes from SD3 Medium, NOVA~\citep{nova} from its
own from-scratch text-to-image model, and GameNGen~\citep{gamengen} and
ART$\cdot$V~\citep{artv} from Stable Diffusion. Our setting belongs to this
group, and Pyramid Flow is the closest published point of comparison in
Table~\ref{tab:scaling}. 

\section{Implementation Details}
\label{app:impl}

\paragraph{Optimization} All models are trained with AdamW~\citep{loshchilov2019adamw} ($\beta_1{=}0.9$, $\beta_2{=}0.999$,
$\epsilon{=}10^{-10}$, weight decay $0$) under a constant learning-rate schedule with a linear warmup of
$500$ steps per data-parallel replica. The learning rate is $3.5\times10^{-5}$, except for the final SFT stage of the scaled model
(Sec.~\ref{sec:scaling}), which uses $1\times10^{-5}$ with all other settings unchanged. Gradients are clipped to a norm of $0.5$.
Training uses bf16 mixed precision~\citep{kalamkar2019bfloat16,micikevicius2018mixed} on an H20 cluster with a
global batch size of $128$. We do not use an EMA of the weights.

\paragraph{Objective} We use flow matching with a linear schedule~\citep{lipman2023flow,liu2023rectifiedflow}. Timesteps are drawn from a
logit-normal distribution ($\mu{=}0$, $\sigma{=}1$) and warped by a training
shift of $3.0$. Each latent
position receives an independently sampled timestep and independent noise; when the chunk size exceeds
one, a single timestep is shared within a chunk and sampled independently across chunks. Text conditions are dropped with
probability $0.1$ during training to enable classifier-free guidance at
inference~\citep{ho2022cfg}.

\paragraph{Sampling} We sample with a flow-adapted DPM-Solver++ multistep
implementation~\citep{lu2025dpmsolverpp}, using $50$
steps and an inference timestep shift of $7.0$ (versus $3.0$ during training) in all experiments.
Classifier-free guidance uses an empty-string negative prompt and no separate guidance on the history
stream; its text scale is $6.5$ for all controlled ablations, diagnostics, and visualizations
(Secs.~\ref{sec:setup}--\ref{sec:image-vae} and the corresponding appendices), and $9.5$ for the
scaling experiments (Sec.~\ref{sec:scaling}). Videos have $81$ frames with the video VAE and $80$ frames with the image VAE,
all at $16$\,fps, and are generated at 256p in the controlled
experiments and at $480\times832$ in the scaling experiments. Each latent
position covers four frames: it holds one latent of the video VAE (except the
first, which covers a single frame) or four per-frame latents of the image VAE
merged by the temporal down layer. The controlled causal models generate $20$ chunks,
with an extra first-frame chunk for the video VAE; the scaled model generates $5$ chunks of
four latent positions each.

\subsection{Architecture}
\label{app:arch}

The DiT is warm-started from an adapted 1.6B-parameter SANA-1.5 checkpoint~\citep{sana15} and keeps its width and depth:
$20$ blocks at hidden width $2240$, $20$ attention heads of dimension $112$. The input layer is a 3D patch embedding with patch size $(1,2,2)$ over the
latent, i.e.\ no temporal patching at this stage; temporal compression is handled separately
(Appendix~\ref{app:patch}). Captions are encoded by a frozen
Gemma-2-2B-IT decoder~\citep{gemma2} at a maximum length of $512$ tokens.

\paragraph{Temporal Convolution in the MLP} Following SANA-Video~\citep{sanavideo}, each block's MLP carries a zero-initialized temporal
convolution with kernel size 3 and a shortcut after the spatial GLU path. In our
implementation this is a full (not depthwise) $\mathrm{Conv2d}$ with kernel
$(3,1)$ over the $(T, HW)$ layout. 

\paragraph{Attention} Following SANA-Video~\citep{sanavideo}, we apply a ReLU feature map to $q$ and $k$ and only then apply
RoPE~\citep{su2021roformer} to the features, making the two compatible.
Positions use axis-wise video RoPE; we set $\theta{=}10^4$ along time and
$\theta{=}2\times10^3$ along height and width.
We replace $25\%$ of the
blocks with softmax attention; with depth $20$ the rule places softmax at blocks
$\{3, 7, 11, 15, 19\}$, i.e.\ every fourth block ending at the last.

\subsection{Temporal Patch Layer}
\label{app:patch}

With the image VAE every latent frame corresponds to one video frame, so we restore the token budget
inside the DiT rather than in the VAE. Immediately after the patch embedding, a down-projection maps
each group of $p{=}4$ adjacent latent frames to a single token group, and a matching up-projection
inverts it before the output layer; the transformer blocks therefore operate on $80/4 = 20$ temporal
positions.

Both projections are implemented as a strided convolution along time with kernel size equal to stride,
which is exactly a bias-free linear map on the $p$ concatenated frames. The pair costs
$40$M parameters, about $2\%$ of the model; Fig.~\ref{fig:crp}(c) shows where
they sit.

\subsection{Adapting SANA-1.5 to the Wan VAE Latent Space}
\label{app:vae-adapt}

SANA-1.5 is pretrained in the DC-AE latent space~\citep{dcae}, whose $32\times$
spatial compression and $32$ latent channels differ from the Wan2.1 VAE's
$8\times$ and $16$, so the pretrained input and output layers cannot be reused as
they are. We adapt them in two stages, following the embedding-alignment recipe
of DC-VideoGen~\citep{dcvideogen}.

The first stage aligns the two embedding spaces. The new input layer is a
randomly initialized 3D patch embedding with stride $(1,2,2)$ over the Wan
latent, and its output tokens are average-pooled $2\times2$ so that the grid
matches the one the original SANA-1.5 input layer produces from the DC-AE latent
of the same image. With everything else frozen, we train the $146$K parameters of
this layer to minimize the MSE between the two embeddings. It converges quickly:
the cosine similarity between them exceeds $0.4$ within $500$ steps and then climbs to about $0.7$.
The second stage freezes the DiT blocks and trains only the input and output
layers with the flow-matching loss on Wan VAE latents, under the same prediction
parameterization as the run it initializes. This stage is image-only and
consumes on the order of 1M samples.

The DC-VideoGen code had not yet been released at the time of writing, so the
procedure above is our own reimplementation from the paper's description. It
sped up convergence of the subsequent video training in our runs.

\section{Training Data}
\label{app:data}

Our training data come from an internal video corpus. Raw videos are first split
into single-shot clips with a shot-boundary detector, and the starting frame of
each clip is chosen for sharpness; the clips are then deduplicated and
concept-balanced using video-level embeddings. A cascade of filters discards
clips with low aesthetic or technical quality, visual blur, and little or no
motion as measured by optical flow. The experiments in this paper use a subset of about 15M clips drawn from
the resulting pool. Each clip is 5--6 seconds long, and during training we
uniformly sample a 5-second window at 16 fps.

Captions are generated for whole clips by a vision-language model and come in
short, medium, and long variants, which are sampled during training with
probabilities of 10\%, 30\%, and 60\%, respectively. The controlled experiments
and the subsequent 256p/480p scaling experiments use the same dataset and differ
only in training resolution. After these stages, the scaled model is further
fine-tuned (SFT) on 0.5M internal high-quality videos; the
controlled experiments do not include this stage.

\section{Evaluation Protocol}
\label{app:eval}

\subsection{VBench}
\label{app:eval-vbench}

We use the official VBench T2V evaluation~\citep{huang2024vbench}. For the
scaled model, following the evaluation protocol in the official VBench
repository, we generate videos with the augmented prompts from
\citet{selfforcing} but compute scores using the original short VBench prompts.
The results for Optica are measured at \(480\times832\), 80 frames, and
16 fps. We sample for 50 steps with a flow-adapted DPM-Solver++
implementation~\citep{lu2025dpmsolverpp}, using a text CFG scale of 9.5 and an
inference flow shift of 7.0. The controlled ablations (Table~\ref{tab:ladder}) instead generate 256p
videos directly from the standard short VBench prompts, with 50 steps, a flow shift of
7.0, and a text CFG scale of 6.5 (Appendix~\ref{app:impl}).

In Table~\ref{tab:scaling}, the scores of Step-Video, Open-Sora-2.0,
HunyuanVideo, Wan2.1-14B, CogVideoX1.5, and LTX-Video are taken from
the public VBench leaderboard~\citep{vbenchleaderboard}; those of Wan2.1-1.3B,
SkyReels-V2, MAGI-1, and CausVid from Table~1 of \citet{selfforcing}, which evaluates CausVid
through its official Wan2.1-1.3B implementation; and all other rows from their
own papers. The Rolling Forcing Total is taken from
\citet{contextforcing}, since \citet{rollingforcing} reports only Quality and
Semantic. Evaluation protocols, including prompt rewriting, resolution, and
sampling, differ across rows.

\subsection{Per-Frame Quality and the History-Corruption Diagnostic}
\label{app:eval-quality}

To quantify quality degradation during autoregressive rollouts, we compare the
per-frame image quality of teacher forcing and autoregressive generation on
1,000 real video clips. The autoregressive setting performs a complete
autoregressive rollout. The teacher-forcing setting instead uses real latents as
history and generates each chunk sequentially. Both
settings use a flow shift of 7.0 and a text CFG scale of 6.5.
For the video VAE, $81$ frames correspond to $21$ latent positions;
Fig.~\ref{fig:history-quality} reports the first $20$ positions
($0$--$19$), including the initial position.

We use the public Q-ReAlign Mini checkpoint to evaluate
the image quality of every frame. Q-ReAlign reuses the discrete text-defined
rating formulation introduced by Q-Align~\citep{wu2024qalign}. Specifically,
the model predicts probabilities for five ratings (excellent, good, fair,
poor, and bad), which are weighted by 1.00, 0.75, 0.50, 0.25, and 0.00,
respectively, to obtain the final score. We first average all frames within each
chunk, and then use each video clip as the statistical unit when computing the
mean and standard error.

To more directly measure model sensitivity to erroneous history, we corrupt
only the ground-truth history used as the condition while leaving the current
chunk to be generated unchanged. We use five corruption types: brightness,
contrast, saturation, elastic transform, and smooth grid warp. The magnitude of
each of the first three corruptions is 0.20. Elastic transform uses severity 3
from the imagecorruptions library~\citep{michaelis2019benchmarking}, which
implements the corruptions of \citet{hendrycks2019robustness}, and the maximum
displacement of smooth grid warp is 0.03 times the shorter image side. Each
corruption is triggered independently on every frame with probability 0.3, so
multiple corruptions may be applied to the same frame, while a frame on which
none is triggered remains unchanged.
In Fig.~\ref{fig:history-quality}, \(\Delta Q\) denotes
\(Q_{\mathrm{corrupted}}-Q_{\mathrm{clean}}\) for the same video clip and
chunk. All curves report the mean and standard error over 1,000 clips.

\subsection{VAE Decoder Perturbation Analysis}
\label{app:eval-relay}

To analyze the history dependence introduced by the video VAE, we perform
latent-perturbation experiments directly, without involving the DiT. We first
encode 1,000 real video clips using the Wan2.1 VAE, then add Gaussian
perturbations to one latent frame:
\[
z_i' = z_i + 3\epsilon,\qquad \epsilon\sim\mathcal{N}(0,I),
\]
while leaving all other latent frames unchanged. We then decode the original
and perturbed latents separately and compute the pixel-space RMS for every
output frame. This metric first computes the root mean square over the spatial
dimensions and then averages over channels.
Figure~\ref{fig:vae-decoder-relay}(a) reports the mean over all clips, with
shading indicating the 95\% confidence interval computed by clip-level
bootstrap.

For Fig.~\ref{fig:vae-decoder-relay}(b), at each latent position we pool all
clips and spatial positions, estimate the mean and variance of each channel,
and compute \(\mathrm{KL}(q(z_i)\,\|\,\mathcal{N}(0,I))\) using a
moment-matched Gaussian. We finally average over channels to obtain the KL per
latent dimension.

\section{History Dependence in the Video VAE}
\label{app:video-vae}

Like many common video VAEs, the Wan2.1 VAE we adopt~\citep{wan2025wan} uses an
architecture based on causal temporal convolutions; related temporally
compressive representations are used throughout latent video generation~\citep{cogvideox,kong2024hunyuanvideo,ltxvideo,magvit,magvitv2,cvvae,tats}.
Related consequences of this design have been noted before, though not for
autoregressive rollouts: IV-VAE~\citep{ivvae} shows that causal temporal
convolution gives frames unequal information access and unbalanced
reconstruction quality, and UniTemp~\citep{unitemp} identifies the same
temporal bias as an obstacle to generating in reverse temporal order. Neither
concerns exposure bias or error accumulation during autoregressive generation,
which is the question our analysis targets. We find that, when paired with the VAE
training objective, the causal architecture creates dependencies between
earlier and later frames. Specifically, perturbing an earlier frame affects the
decoded results at later positions. Per-frame latents, which avoid this coupling, have
precedent: early latent video diffusion models operated on them~\citep{blattmann2023videoldm,blattmann2023svd}, and FAR~\citep{far} adopts an
image VAE for frame-level autoregression, though without arguing for the
choice.

\begin{figure*}[!ht]
\centering
\includegraphics[width=0.75\textwidth]{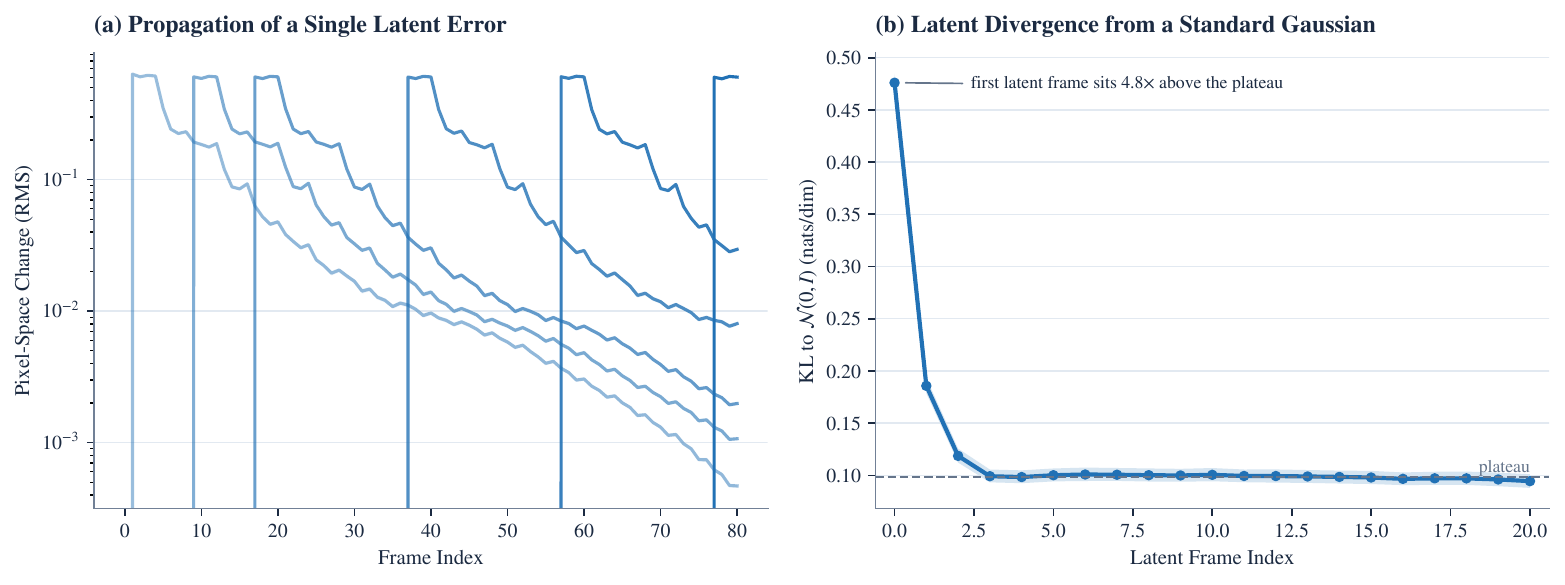}
  \caption{\textbf{The video VAE couples latent frames.} 
  \textbf{(a)} Pixel-space RMS change over output-frame index after perturbing a
  single latent frame and decoding the same sequence, for several perturbation
  locations; every perturbation produces a persistent tail that reaches the end
  of the clip.
  \textbf{(b)} KL between per-channel moment-matched Gaussian approximations
  and the standard Gaussian prior, averaged over channels at each latent
  position. The KL is largest at the first latent frame and lower from
  latent frame 3 onward.}
\label{fig:vae-decoder-relay}
\end{figure*}

As shown in Fig.~\ref{fig:vae-decoder-relay}(a), an
error introduced at one position propagates to subsequent positions. Under the
common setting of 5 seconds at 16 fps, an error introduced in an
early latent propagates all the way to the last frame. We also examine the KL
divergence between per-channel moment-matched Gaussian approximations of
the Wan VAE latents and the standard Gaussian prior
(Fig.~\ref{fig:vae-decoder-relay}(b)).

Under this moment-matched Gaussian approximation, the first latent frame
has a substantially larger KL to the prior than later frames. From latent frame 3 onward, the KL stays on a plateau
4.8\(\times\) below that of the first latent frame. The
first latent thus has different statistics from the later ones, matching the
two latent types of Sec.~\ref{sec:vae}. For causal video, the propagation in
Fig.~\ref{fig:vae-decoder-relay}(a) means that once a signal at an earlier
position is generated incorrectly, the VAE continually carries it into all
subsequent positions.

Note that we remove all history dependence from the VAE, not only the
unnecessary part, and leave history modeling entirely to the DiT, where CRP
handles it. Another
possible path is to retain necessary dependence on history in the VAE while
removing only unnecessary dependence. This alternative could further compress
the signal and reduce the modeling difficulty of the DiT, thereby improving
overall performance. We discuss this possibility in the hope of motivating
further exploration by the community.

\section{VAE Reconstruction: Image vs.\ Video Mode}
\label{app:vae-recon}

We verify that changing the temporal mode of the VAE does not improve the
downstream model by sacrificing reconstruction fidelity. Both settings use the
same pretrained Wan2.1 VAE checkpoint; video mode retains its causal temporal
convolutions, whereas image mode encodes and decodes every frame independently.
We evaluate both modes on the same 1,000 real-video clips, each containing 81
frames at \(192\times336\) resolution. For each mode, we encode and immediately
decode the clip without involving the DiT.

\begin{table}[!ht]
  \caption{Reconstruction fidelity of the same Wan2.1 VAE checkpoint under
  video-mode and image-mode encoding/decoding. Values are mean per-frame PSNR
  \(\pm\) standard error over 1,000 paired real-video clips. Encoding latency is
  the synchronized batch-one mean for an 81-frame clip.}
  \label{tab:vae-reconstruction}
  \begin{center}
  \begin{tabular}{lcc}
  \toprule
  {VAE mode} & {PSNR (dB)} & {Encode (ms/clip)} \\
  \midrule
  Video mode (causal temporal) & \(27.93 \pm 0.13\) & \(754\) \\
  Image mode (per-frame)       & \(29.04 \pm 0.15\) & \(909\) \\
  \bottomrule
  \end{tabular}
  \end{center}
  \end{table}

We clamp both the input and reconstruction to \([0,1]\), compute PSNR for every frame, average frames within each clip, and finally average over
clips. Uncertainty is computed with clips as the statistical unit.

\begin{figure*}[!ht]
  \centering
  \includegraphics[width=\textwidth]{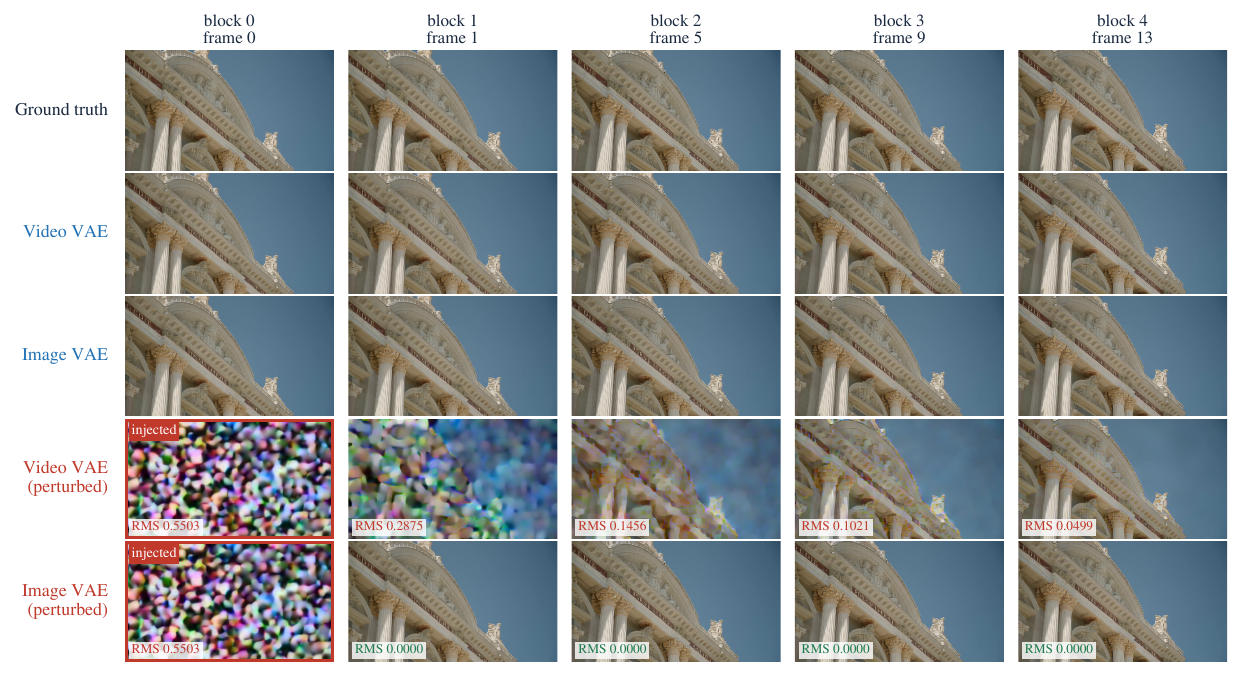}
  \caption{\textbf{Reconstruction quality is comparable between the two VAE
  modes, but only the video VAE propagates a latent error across frames.} No DiT
  is involved. Columns are the pixel frames $0,1,5,9,13$. Rows $2$--$3$ are ordinary
  reconstructions. Rows $4$--$5$ repeat them after adding $3\epsilon$, $\epsilon\sim\mathcal{N}(0,I)$, only to
  the latent frame that owns pixel frame~$0$; the red box marks that injected
  column, which therefore changes in both modes. Insets give the pixel-space RMS
  change from the corresponding clean reconstruction. Past the injected column
  the video VAE still drifts by $0.05$--$0.29$ RMS, while the image VAE is
  exactly $0$.}
  \label{fig:vae-recon-qualitative}
  \end{figure*}
Image mode improves paired PSNR by \(1.11\pm0.04\) dB. Figure~\ref{fig:vae-recon-qualitative} shows the corresponding qualitative
comparison, and additionally visualizes in pixel space the latent coupling
measured in Fig.~\ref{fig:vae-decoder-relay}(a). Clean reconstructions from the
two modes are visually close, consistent with the small paired PSNR gap in
Table~\ref{tab:vae-reconstruction}. The perturbed rows behave very differently.
Following the protocol of Appendix~\ref{app:eval-relay}, we perturb the single
latent frame that owns pixel frame~0 and decode the sequence again. In video
mode, the corruption remains plainly visible at every
subsequent displayed frame. In image mode, the same latent frame owns pixel
frame~0 alone, and every later frame is recovered bit-for-bit.

\end{document}